\documentclass{IEEEtran}
\usepackage{cite}
\usepackage{amsmath,amssymb,amsfonts}
\usepackage{graphicx}
\usepackage{textcomp,nicefrac}
\def\BibTeX{{\rm B\kern-.05em{\sc i\kern-.025em b}\kern-.08em
T\kern-.1667em\lower.7ex\hbox{E}\kern-.125emX}}

\usepackage{array}
\usepackage[hyphens]{url}
\usepackage{hyperref}
\usepackage[all]{hypcap}
\usepackage{siunitx}
\usepackage{pdfpages}
\usepackage[caption=false,font=footnotesize]{subfig}


\makeatletter
\newcommand\Autoref[1]{\@first@ref#1,@}
\def\@throw@dot#1.#2@{#1}
\def\@set@refname#1{
    \edef\@tmp{\getrefbykeydefault{#1}{anchor}{}}%
    \xdef\@tmp{\expandafter\@throw@dot\@tmp.@}%
    \ltx@IfUndefined{\@tmp autorefnameplural}%
         {\def\@refname{\@nameuse{\@tmp autorefname}s}}%
         {\def\@refname{\@nameuse{\@tmp autorefnameplural}}}%
}
\def\@first@ref#1,#2{%
  \ifx#2@\autoref{#1}\let\@nextref\@gobble
  \else%
    \@set@refname{#1}
    \@refname~\ref{#1}
    \let\@nextref\@next@ref
  \fi%
  \@nextref#2%
}
\def\@next@ref#1,#2{%
   \ifx#2@ and~\ref{#1}\let\@nextref\@gobble
   \else, \ref{#1}
   \fi%
   \@nextref#2%
}
\makeatother

\begin{document}
\title{Neural Network Acceleration on MPSoC board: Integrating SLAC’s SNL, Rogue Software and Auto-SNL}
\author{
Hamza Ezzaoui Rahali, Abhilasha Dave, Larry Ruckman, Mohammad Mehdi Rahimifar, Audrey C. Therrien, James J. Russel, Ryan T. Herbst
\thanks{This work has been submitted to the \textsc{IEEE Transactions on Nuclear Science} for possible publication. Copyright may be transferred without notice, after which this version may no longer be accessible.}
\thanks{Authors A. D., L. R., J. R., R. H. are with the SLAC National Accelerator Laboratory, 2575 Sand Hill Road, Menlo Park, CA 94025, USA (e-mails: \{adave, russell, ruckman, rherbst\}@slac.stanford.edu).}
\thanks{Authors H. E. R., A. C. T. are with the Interdisciplinary Institute for Technological Innovation, 3000 University Blvd, Sherbrooke, QC J1K 0A5, Canada. Author M. M. R. graduated from the University of Sherbrooke, 2500 University Blvd, Sherbrooke, QC J1N 3C6, Canada (emails: \{hamza.rahali, mohammad.mehdi.rahimifar, audrey.corbeil.therrien\}@usherbrooke.ca).}
}

\maketitle

\begin{abstract}
The LCLS-II Free Electron Laser (FEL) will generate X-ray pulses for beamline experiments at rates of up to 1~MHz, with detectors producing data throughputs exceeding 1 TB/s. Managing such massive data streams presents significant challenges, as transmission and storage infrastructures become prohibitively expensive. Machine learning (ML) offers a promising solution for real-time data reduction, but conventional implementations introduce excessive latency, making them unsuitable for high-speed experimental environments. To address these challenges, SLAC developed the SLAC Neural Network Library (SNL), a specialized framework designed to deploy real-time ML inference models on Field-Programmable Gate Arrays (FPGA). SNL's key feature is the ability to dynamically update model weights without requiring FPGA resynthesis, enhancing flexibility for adaptive learning applications. To further enhance usability and accessibility, we introduce Auto-SNL, a Python extension that streamlines the process of converting Python-based neural network models into SNL-compatible high-level synthesis code. This paper presents a benchmark comparison against hls4ml, the current state-of-the-art tool, across multiple neural network architectures, fixed-point precisions, and synthesis configurations targeting a Xilinx ZCU102 FPGA. The results showed that SNL achieves competitive or superior latency in most tested architectures, while in some cases also offering FPGA resource savings. This adaptation demonstrates SNL's versatility, opening new opportunities for researchers and academics in fields such as high-energy physics, medical imaging, robotics, and many more.
\end{abstract}


\section{Introduction}
\label{sec:introduction}
Modern ultra-high-rate (UHR) experimental facilities such as the Linac Coherent Light Source II (LCLS‑II) push the limits of scientific imaging by delivering X-ray pulses at repetition rates up to 1~MHz. These unprecedented rates allow for detailed time-resolved studies in material science, chemistry, and biology, but also generate data at a staggering rate of over 1~TB/s, far outpacing the capacity of conventional data storage and processing systems~\cite{herbst2022implementation}.

Machine learning (ML) has emerged as a powerful approach for real-time data reduction by moving inference to the edge~\cite{duarte_2019, rahimifar_2023}. While CPUs and GPUs are common ML deployment platforms, they often fall short in meeting the ultra-low-latency demands of facilities like LCLS-II. In such environments, inference pipelines must operate within microsecond-scale latency budgets, making FPGAs at the edge a compelling choice for low-latency, high-throughput ML inference.

In response, SLAC developed the SLAC Neural Network Library (SNL)~\cite{herbst2022implementation}, a domain-specific HLS-based framework designed to deploy ML models on FPGAs for real-time experimental applications. A defining feature of SNL is its ability to dynamically reload neural network (NN) weights and biases without requiring FPGA resynthesis, thus supporting rapid iteration, retraining, and deployment in mission-critical environments~\cite{dave2025fpga}. To further democratize FPGA-based ML deployment, we introduce Auto-SNL, a Python extension that automates the conversion of Python-defined models into SNL-compatible high-level synthesis (HLS) code. Auto-SNL lowers the barrier to entry by abstracting away hardware intricacies, allowing non-expert users to generate optimized bitfiles and control hardware parameters without directly engaging with FPGA toolchains.

One of the main goals of this work is to position SNL’s current capabilities within the broader landscape of ML-to-FPGA frameworks. While SNL and Auto-SNL address the specific needs of SLAC’s ongoing experiments, it is important to understand how they compare with other available tools in terms of features, flexibility, design-space exploration, and synthesis efficiency.

Several FPGA-oriented ML deployment frameworks have also emerged in recent years. Vendor-supported toolchains such as Xilinx Vitis AI~\cite{vitisai_doc} and Intel OpenVINO~\cite{intel_openvino} provide optimized pipelines for neural networks using pre-built and proprietary IP blocks. FINN~\cite{blott2018finn} targets extreme quantization for ultra-low latency embedded inference, while other research projects explore domain-specific accelerators or specialize in a particular set of ML architectures~\cite{chen2019eyeriss, gao2020edgedrnn}.

To evaluate SNL, we selected hls4ml~\cite{Duarte:2018ite} as the open-source counterpart in our comparative study. This mature HLS toolchain supports a broad range of network architectures, is actively maintained, and has been adopted in multiple scientific computing domains~\cite{summers2020fast, khoda2023ultra}. Unlike SNL, hls4ml requires network weights and biases to be embedded into the FPGA fabric during synthesis. This limits runtime flexibility, but the framework compensates by offering user-configurable synthesis parameters, such as the strategy, enabling latency- or resource-optimized implementations, and the reuse factor, controlling the number of times a multiplier unit is reused during inference~\cite{Aarrestad:2021zos}. These trade-offs play a crucial role in applications where design constraints vary widely. As such, hls4ml serves as a well-established and representative comparison point, keeping our study informative yet tractable in size.

The comparison is structured as a benchmark measuring both inference latency and FPGA resource utilization across multiple network architectures and, when available, different hardware synthesis settings. In designing this benchmark, we account for multiple factors that critically impact performance, including data-flow patterns, synthesis strategies, and model complexity. These considerations, previously explored in earlier work~\cite{jia2024analysis}, frame the trade-offs faced by real-time FPGA-based ML systems and motivate the extended analysis presented in this work. The benchmark spans both fully-connected neural networks (FCNNs) and convolutional neural networks (CNNs) from physics, audio, and vision domains. Results are reported in terms of logic, DSP, and BRAM consumption, as well as absolute latency~(\SI{}{\micro\second}), targeting a Zynq UltraScale+ MPSoC ZCU102 FPGA~\cite{amd2024zcu102}, with reports and designs generated using Vitis HLS and Vivado tools.

In what follows, we first provide background on both frameworks and their respective workflows on the ZCU102 board: \Autoref{sec:snl-flow} describes SNL, Auto-SNL, and the integration with Rogue software on the MPSoC platform, whereas~\Autoref{sec:hls4ml-flow} introduces hls4ml and explains its corresponding workflow on the ZCU102. These preliminary sections are followed by~\Autoref{sec:benchmarking}, where we define the benchmarking protocol, including the models and the synthesis parameter design space. In \Autoref{sec:results} and \Autoref{sec:discussion}, we present and analyze the comparative results between SNL and hls4ml across different hardware parameters, highlighting trade-offs in latency and resource utilization. Finally, \Autoref{sec:conclusion} summarizes the key outcomes of this work, discussing limitations and directions for future work.

\section{SNL and Auto-SNL Workflow with SLAC’s Rogue Software}
\label{sec:snl-flow}

SNL is a high-level synthesis framework capable of deploying NNs into the programmable logic (PL) of FPGAs, with optimizations for ultra-low-latency inference in resource-constrained environments. SNL’s suitability for low-latency inference in demanding experiments has been demonstrated through its FPGA implementation of SpeckleNN~\cite{dave2025fpga}, a deep learning model designed for real-time X-ray Single-Particle Imaging (SPI) at XFEL facilities. This application helps showcase SNL’s ability to execute scientific ML workloads close to the data source, reducing transfer and processing delays at the earliest stage of detector acquisition. Another key feature of SNL is its ability to dynamically update model weights and biases without FPGA re-synthesis, allowing models to be reconfigured and retrained on the fly. This is particularly advantageous for adaptive scientific experiments, where frequent updates and fine-tuning are required without interrupting real-time processing.

\begin{figure}[!t]
    \centering
    \includegraphics[width=\linewidth]{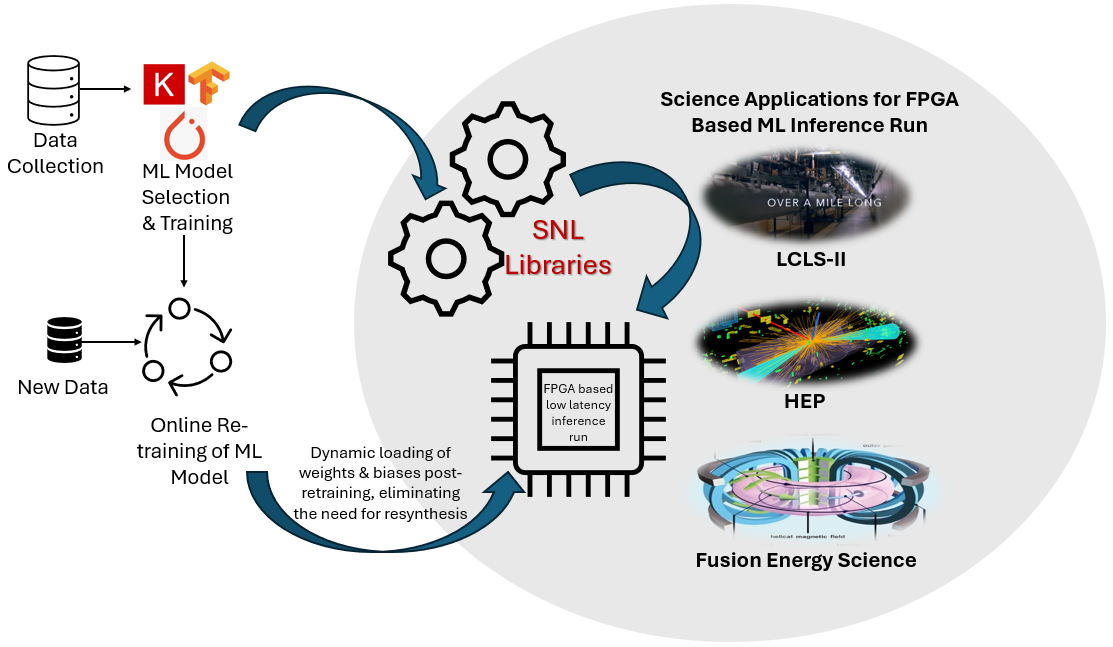}
    \caption{High-level view of SNL's workflow.}
    \label{fig:SNL_flow}
\end{figure}

\begin{figure*}
    \centering
    \includegraphics[width=\linewidth]{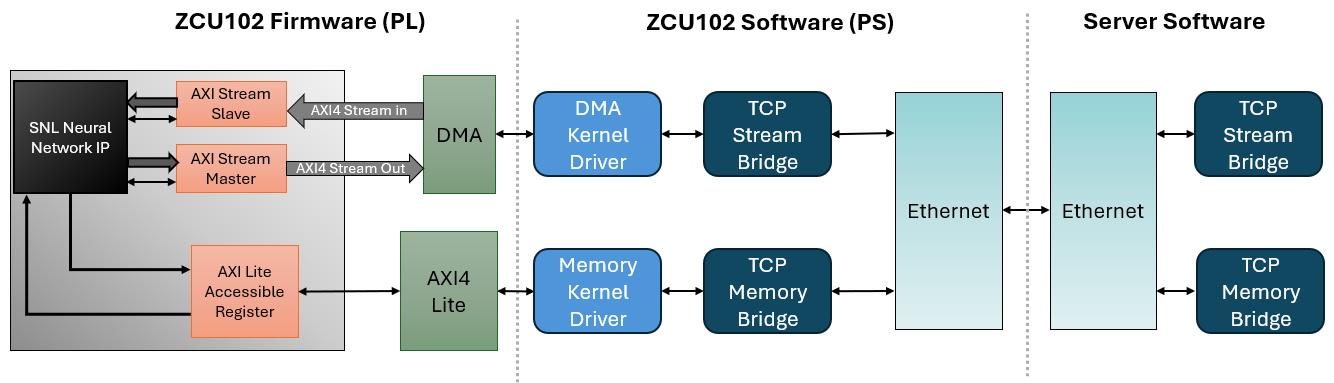}
    \caption{SNL's workflow for NN Deployment on
ZCU102: Hardware (PL) -- Software (SNL) -- Rogue design flow.}
    \label{fig:ZCU102-design-flow}
\end{figure*}

As shown through~\Autoref{fig:SNL_flow, fig:ZCU102-design-flow}, SNL deploys the NN layers entirely in PL for maximum efficiency. Weights and biases are loaded via AXI-Lite registers, while AXI-Stream facilitates real-time input and output data flow during inference. The direct memory access (DMA) engine ensures high-speed data transfer between processing elements, minimizing latency and optimizing performance.

\begin{figure}[!t]
    \centering
    \includegraphics[width=\linewidth]{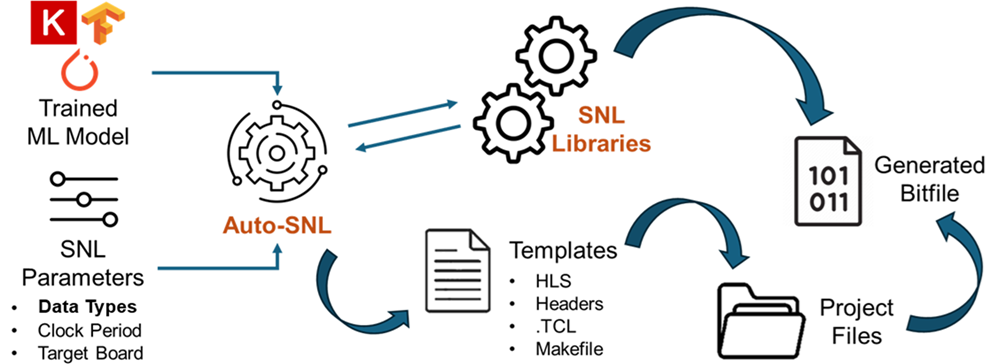}
    \caption{Auto-SNL conversion and implementation workflow.}
    \label{fig:Auto-SNL_flow}
\end{figure}

The ZCU102 platform, running a custom PetaLinux image, manages device drivers, while SLAC’s Rogue software provides configurable hardware interaction~\cite{rogue2025}. This architecture, described in~\Autoref{fig:ZCU102-design-flow}, enables a streamlined, hardware-optimized pipeline for deploying neural networks on FPGAs, ensuring efficient real-time inference for scientific applications. A TCP stream bridge converts input and output data into TCP packets for Ethernet transmission, enabling seamless interaction with run control software.

To make the deployment of NN models on FPGA more accessible, we developed Auto-SNL, a complementary tool that simplifies the conversion of trained ML models into SNL-based HLS code. Auto-SNL bridges the gap between high-level frameworks like Keras and TensorFlow and SNL’s hardware-focused architecture. As illustrated in~\Autoref{fig:Auto-SNL_flow}, Auto-SNL requires a trained ML model and parameters such as data types, clock period, and the target FPGA board. Using these inputs, Auto-SNL writes the necessary files for building the SNL project on supported hardware. The generated project is then implemented to produce the bitfile, which can be integrated with the custom PetaLinux image and deployed on the FPGA. Throughout this process, Auto-SNL ensures compatibility with SNL and optimizes the generated code for the intended hardware platform.

One of the core objectives of Auto-SNL is to offer users the ability to adjust hardware parameters without requiring an in-depth understanding of the underlying HLS implementation. This allows fine-tuning of ML implementations, making it possible to tailor the deployment to specific requirements without manually modifying the generated code. Additionally, Auto-SNL aims to continue SNL’s modular design by providing a flexible and extensible workflow. Support for new hardware platforms and ML frameworks can be added by defining appropriate templates, ensuring that this tool adapts to evolving FPGA technologies and ML frameworks, and enabling rapid prototyping and deployment of ML inference systems.

\section{Workflow for deployment and inference on the ZCU102 using hls4ml}
\label{sec:hls4ml-flow}

\begin{figure*}
    \centering
    \includegraphics[width=\linewidth]{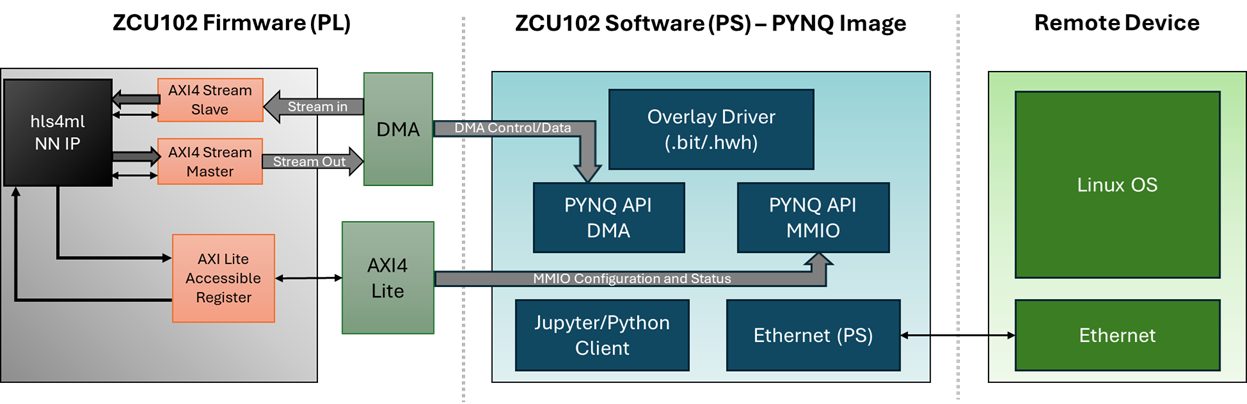}
    \caption{Standard hls4ml workflow for a streaming-based NN deployment on a ZCU102 running a PYNQ image.}
    \label{fig:zcu102-flow-hls4ml}
\end{figure*}

\Autoref{fig:zcu102-flow-hls4ml} shows a standard hls4ml workflow for deploying NNs on the ZCU102 board configured with a PYNQ image~\cite{pynq_doc}. Starting with a high-level model description (Keras, Pytorch, or ONNX) and a hardware-related configuration, hls4ml generates synthesizable C++ code, which is then packaged as a custom hardware IP core implementing the neural network in the PL. 

During deployment, the custom IP connects to the Zynq processing system (PS) via AXI interfaces. An AXI-Lite interface provides control and status registers, while data movement can depend on hls4ml's \textit{I/O type} parameter. For \textit{IO Stream}, the option matching SNL's streaming interface, the IP core exchanges data with the PS through AXI-Stream, typically coupled with DMA engines to sustain high-throughput transfers between PS memory and the accelerator. Unlike SNL, where weights and biases can be updated at runtime, hls4ml embeds all model parameters into the PL during synthesis, meaning any retraining or parameter update requires regenerating and reloading a new bitstream.

Once the block design is implemented, the toolchain generates a bitstream file (.bit), defining the PL layout, including the NN accelerator, its interfacing logic, the embedded model parameters, and any other auxiliary IP. Alongside the bitstream, a hardware handoff file (.hwh) is produced, containing a machine-readable description of the block design, address map, and register definitions. On the PYNQ Linux OS running on the PS, these files are loaded at runtime through the Overlay API: the .bit configures the PL, while the .hwh enables automatic generation of Python-accessible drivers for controlling the accelerator and initiating data transfers.

At runtime, inference can be orchestrated from the PS, either through local Jupyter notebooks or remotely over Ethernet, with inputs and outputs exchanged using the data path defined at synthesis. This separation between hardware generation and runtime control allows hls4ml to combine FPGA-level acceleration with the flexibility of high-level Python interfaces, while giving developers control over a wide range of synthesis parameters, enabling fine-grained command over the latency-resource design space to suit diverse application requirements.

\section{Benchmarking}
\label{sec:benchmarking}

To evaluate the performance and workflow characteristics of SNL, we conducted a comparative synthesis benchmark against the widely adopted hls4ml toolchain. The motivation for this study is twofold. First, the side-by-side evaluation enables a comprehensive assessment of latency and resource utilization across the two toolchains. Second, the results provide insight into where SNL's current capabilities align with, or diverge from, established ML-to-FPGA workflows, thereby informing both user decision-making and potential directions for future development.

Our benchmark comprises four neural networks representing both domain-specific and general-purpose inference workloads. The selection includes (1) a particle jet classifier for high-energy physics applications~\cite{hls4ml_jet}, along with three models from the MLPerf Tiny benchmark suite~\cite{banbury2021mlperf}: (2) a fully connected autoencoder for anomaly detection in machine operating sounds, (3) a convolutional neural network for keyword spotting (KWS) in audio data, and (4) a convolution-based binary image classification model (VWW). This mix balances diversity in layer types, network depths, and parameter counts, ensuring coverage from compact to moderately complex architectures.

\Autoref{tab:benchmark-desc} summarizes each benchmark item, outlining the dataset used for training and validation, the input size, the intended task, and the performance achieved, measured with either accuracy or area-under-curve (AUC) metrics. All networks were trained on their respective datasets following the pre-processing and input formatting procedures described in the original references~\cite{hls4ml_jet, banbury2021mlperf}. To ensure compatibility with the ZCU102 board and SNL’s current feature set, architectures and input sizes, particularly for the convolutional networks, were adjusted to reduce resource usage and align with SNL's layer support. For classification networks, we also truncate the final softmax activation layer during hardware synthesis. \Autoref{tab:benchmark-arch} illustrates the full network architectures as well as the training hyperparameters.

\begin{table}[htb]
    \footnotesize
    \renewcommand{\arraystretch}{1.5}
    \setlength\tabcolsep{2pt}
    \begin{center}
    \caption{Benchmark task, dataset, and performance summary.}
    \begin{tabular}{c|c|c|c|c}
        \textbf{Model} & \textbf{Dataset} & \textbf{Input size} & \textbf{Task} & \textbf{Performance} \\
        \hline
        Jet & LHC Jet~\cite{dataset_lhcjet} & (16,) & Classification & 74.90\% \\
        Anomaly & ToyADMOS~\cite{dataset_toyadmos} & (320,) & Detection & 0.70 (AUC) \\
        KWS & Speech Commands~\cite{dataset_speechcommands} & (32, 32, 1) & Classification & 59.33\% \\
        VWW & Visual Wake Words~\cite{dataset_visualwakewords} & (49, 10, 1) & Classification & 70.14\% \\
    \end{tabular}
    \label{tab:benchmark-desc}
    \end{center}
\end{table}

In terms of hardware synthesis, our parameter space attempts to capture common design trade-offs in FPGA-based ML acceleration while keeping the experiment tractable. In this context, the first synthesis parameter we considered is \textit{precision}, referring to the bit-width used to represent the inputs and layer weights, which directly impacts resource usage and on-chip numerical accuracy. For both SNL and hls4ml, three representative fixed-point precisions were selected to span high-accuracy, balanced, and resource-efficient operating points. We also varied two additional parameters specific to hls4ml, \textit{strategy} and \textit{reuse factor}. The strategy setting determines whether synthesis prioritizes minimizing latency or conserving FPGA resources. The reuse factor (RF) controls the degree of operator reuse: up to a certain point, higher RF values result in lower parallelism and lower resource use, but increased latency. The complete set of synthesis parameters and their values is summarized in~\Autoref{tab:synthesis-param}.

\begin{table}[htb]
    \footnotesize
    \renewcommand{\arraystretch}{1.5}
    \setlength\tabcolsep{4pt}
    \begin{center}
    \caption{Hardware synthesis parameter space.}
    \begin{tabular}{l|c|c}
        & \textbf{hls4ml} & \textbf{SNL} \\
        \hline
        \textbf{Precision} & $\langle 32, 16 \rangle$, $\langle 16, 6 \rangle$, $\langle 8, 3 \rangle$ & $\langle 32, 16 \rangle$, $\langle 16, 6 \rangle$, $\langle 8, 3 \rangle$ \\
        \textbf{Strategy} & Latency, Resource & N/A \\
        \textbf{Reuse factor} & 1, 2, 4, 8 & N/A \\
        \textbf{IO Type} & Stream & N/A \\
        \textbf{Clock period} & \SI{10}{\nano\second} & \SI{10}{\nano\second} \\
    \end{tabular}
    \label{tab:synthesis-param}
    \end{center}
\end{table}

\begin{table*}
    \footnotesize
    \renewcommand{\arraystretch}{2.5}
    \setlength\tabcolsep{3pt}
    \begin{center}
    \caption{Training hyperparameters and network architectures.}
    \begin{tabular}{c|c|c|c}
        \textbf{Model} & \textbf{Batch size} & \textbf{Learning rate} & \textbf{Architecture} \\
        \hline
        Jet & 1024 & 10\textsuperscript{-4} & $\xrightarrow[\text{Dense}]{\text{Neurons}=64}\ \xrightarrow[\text{ReLU}]{}\ \xrightarrow[\text{Dense}]{32}\ \xrightarrow[\text{ReLU}]{}\ \xrightarrow[\text{Dense}]{32}\ \xrightarrow[\text{ReLU}]{}\ \xrightarrow[\text{Dense}]{5}\ \xrightarrow[\text{Softmax}]{}\ $ \\
        
        Anomaly & 512 & 10\textsuperscript{-4} & $\xrightarrow[\text{Dense}]{16}\ \xrightarrow[\text{ReLU}]{}\ \xrightarrow[\text{Dense}]{32}\ \xrightarrow[\text{ReLU}]{}\ \xrightarrow[\text{Dense}]{32}\ \xrightarrow[\text{ReLU}]{}\ \xrightarrow[\text{Dense}]{8}\ \xrightarrow[\text{ReLU}]{}\ \xrightarrow[\text{Dense}]{32}\ \xrightarrow[\text{ReLU}]{}\ \xrightarrow[\text{Dense}]{32}\ \xrightarrow[\text{ReLU}]{}\ \xrightarrow[\text{Dense}]{16}\ \xrightarrow[\text{ReLU}]{}\ \xrightarrow[\text{Dense}]{320}\ $ \\
        
        KWS & 64 & 10\textsuperscript{-4} & $\xrightarrow[\text{Conv2D}]{\text{Filters}=16\ \text{Kernel}=(5, 5)}\ \xrightarrow[\text{ReLU}]{}\ \xrightarrow[\text{Conv2D}]{8\ (3, 3)}\ \xrightarrow[\text{ReLU}]{}\ \xrightarrow[\text{Dropout}]{\text{Rate}=0.2}\ \xrightarrow[\text{GlobalAveragePooling2D}]{}\ \xrightarrow[\text{Dense}]{12}\ \xrightarrow[\text{Softmax}]{}\ $ \\

        VWW & 32 & 10\textsuperscript{-4} & $\xrightarrow[\text{Conv2D}]{4\ (3, 3)}\ \xrightarrow[\text{ReLU}]{}\ \xrightarrow[\text{AveragePooling2D}]{\text{Pool}=(2, 2)\ \text{Strides}=(2, 2)}\ \xrightarrow[\text{Conv2D}]{4\ (3, 3)}\ \xrightarrow[\text{ReLU}]{}\ \xrightarrow[\text{GlobalAveragePooling2D}]{}\ \xrightarrow[\text{Dense}]{2}\ \xrightarrow[\text{Softmax}]{}\ $
    \end{tabular}
    \label{tab:benchmark-arch}
    \end{center}
\end{table*}

Precision values are denoted as $\langle X, Y \rangle$, where $X$ is the total bit-width and $Y$ is the number of bits representing the signed number above the binary point. In our experiment, we omit RF = 1 in the resource-optimized strategy, following hls4ml's recommendation to avoid this specific combination~\cite{hls4mldocumentation}.

All designs targeted the Xilinx Zynq UltraScale+ MPSoC ZCU102 board using Vitis HLS and Vivado 2023.1~\cite{amd2023Vitis}. For an unbiased and consistent comparison, we set the hls4ml IO type parameter to Stream, matching SNL’s streaming interface. For SNL, the reported latency excludes the one-time streaming transfer of model weights, which are loaded before inference begins. Therefore, for both frameworks, latency is defined as the time between the arrival of the first input at the FPGA’s streaming interface and the production of the corresponding output. Latency results are obtained from the C-synthesis timing analysis, while resource utilization values (BRAM, DSP, FF, LUT) are taken from the post-implementation report.

\section{Results}
\label{sec:results}
We benchmarked multiple neural network architectures using both SNL and hls4ml, running Vitis HLS and Vivado synthesis with the ZCU102 board as the target. The following shows the benchmark results, with synthesis parameters and configurations as described in~\Autoref{sec:benchmarking}.

\begin{figure*}
  \centering
  \begin{minipage}{\linewidth}
    \centering
    \includegraphics[width=\linewidth, page=1]{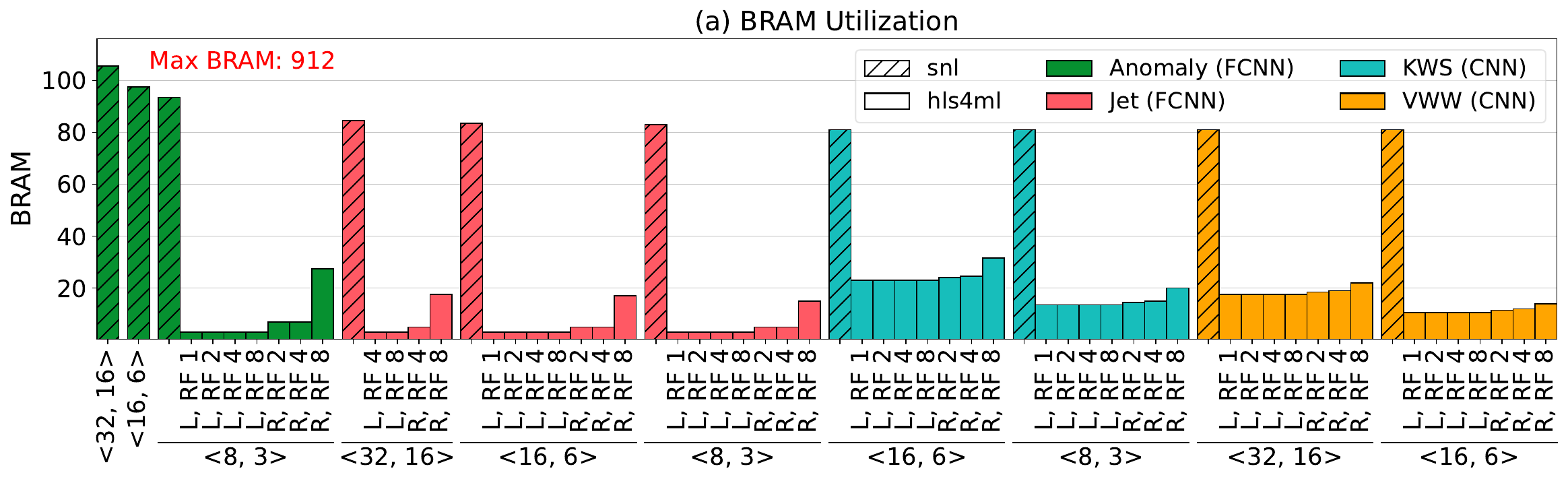}
  \end{minipage}\\[4pt]
  \hfill
  \begin{minipage}{\linewidth}
    \centering
    \includegraphics[width=\linewidth, page=2]{figures/benchmark_bar_plots.pdf}
  \end{minipage}\\[4pt]
  \begin{minipage}{\linewidth}
    \centering
    \includegraphics[width=\linewidth, page=3]{figures/benchmark_bar_plots.pdf}
  \end{minipage}\\[4pt]
  \begin{minipage}{\linewidth}
    \centering
    \includegraphics[width=\linewidth, page=4]{figures/benchmark_bar_plots.pdf}
  \end{minipage}
  \caption{Resource utilization across models and the different synthesis parameters, comparing SNL (bars with shading) and hls4ml. Bars are grouped by model and precision $\langle X, Y \rangle$; hls4ml variants sweep multiple reuse factors and two strategies: Latency (L) and Resource (R).}
  \label{fig:bars-resource}
\end{figure*}

\begin{figure*}
  \centering
  \includegraphics[width=\linewidth, page=5]{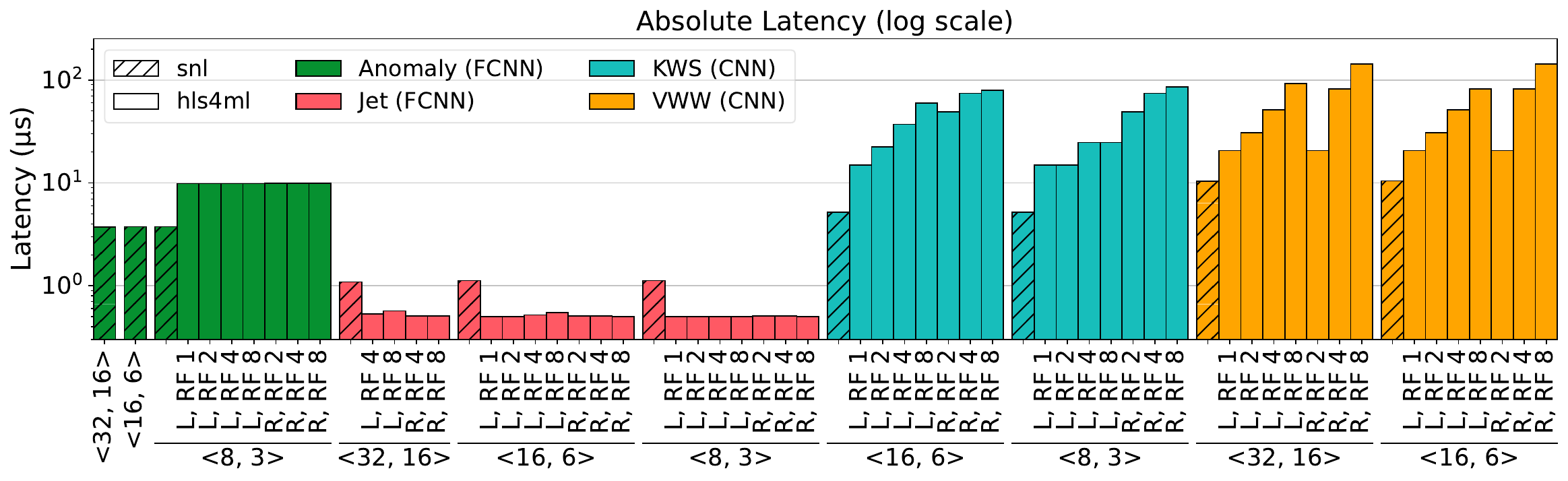}
  \caption{Absolute latency (\SI{}{\micro\second}) across models and the different synthesis parameters, comparing SNL (bars with shading) and hls4ml. Bars are grouped by model and precision $\langle X, Y \rangle$; hls4ml variants sweep multiple reuse factors and two strategies: Latency (L) and Resource (R).}
  \label{fig:bars-latency}
\end{figure*}

\Autoref{fig:bars-resource, fig:bars-latency} present the synthesis results grouped by neural network model and fixed-point precision. Colors indicate the model architecture. For each model–precision pair, the leftmost shaded bar corresponds to the SNL implementation, while the subsequent non-shaded bars show hls4ml implementations ordered by synthesis strategy—latency (L) first, then resource (R)—and by increasing RF. In total, 12 groups are expected (4 architectures, 3 precisions), and each group should contain 1 SNL synthesis followed by 7 hls4ml variants (L/R $\times$ RF).

The vertical axis in~\autoref{fig:bars-resource} reports absolute FPGA resource usage as given by the implementation report: LUT, FF, DSP, and BRAM counts, expressed in absolute numbers. The vertical axis in~\autoref{fig:bars-latency} reports absolute inference latency in microseconds, as reported by C-synthesis. All latency values are device- and clock-specific.

Missing groups in the bar plots indicate synthesis failure of a given model-precision combination with SNL, mainly due to resource limitations. Meanwhile, absent hls4ml bars within the same group reflect hls4ml designs that failed synthesis for the same reason. This layout allows quick comparison between the single SNL design and multiple hls4ml configurations, while making synthesis success and failure immediately apparent.

\section{Discussion}
\label{sec:discussion}
The synthesis results highlight clear trade-offs between SNL and hls4ml, reflecting both the frameworks' differing implementation choices and the latency-resource patterns observed across the model-parameter space.\\
In terms of resource utilization, a high-level view of~\Autoref{fig:bars-resource} shows that SNL consistently uses more BRAM and mostly higher FF counts than hls4ml across models and parameters. Meanwhile, DSP and LUT consumption exhibit an architecture-dependent behavior: for CNN models, SNL typically requires more logic resources than hls4ml, whereas for FCNNs, SNL employs fewer LUTs across all precisions and fewer DSPs in higher ones. Conversely, hls4ml demonstrates a consistent and significant decrease in DSP utilization in lower precisions, particularly with precision $\langle 8, 3 \rangle$. As expected, increasing hls4ml's reuse factor, which determines the number of times the same multiplier unit is reused, consistently lowers DSP usage. LUTs also show a similar trend within the resource-focused strategy.

As noted in~\Autoref{sec:results}, the absence of certain groups in~\Autoref{fig:bars-resource} reflects SNL synthesis failures for specific model-precision combinations. For instance, KWS-$\langle 32, 16 \rangle$ synthesis reports a failure due to exceeding the available ZCU102 DSP and LUT units, while VWW-$\langle 8, 3 \rangle$ failure stems from a framework-related issue.\\
Similarly, missing hls4ml bars within an otherwise present group correspond to configurations that failed. In our benchmark set, hls4ml fails to implement the Anomaly autoencoder for precisions $\langle 32, 16 \rangle$, and $\langle 16, 6 \rangle$, as well as the fully-connected Jet model with precision $\langle 32, 16 \rangle$ and RFs 1 and 2. Post-synthesis reports attribute these failures to exceeding either DSP or LUT availability.

In terms of latency, the trends in~\Autoref{fig:bars-latency}, presented in logarithmic scale, show that SNL achieves lower inference latency than hls4ml in three out of the four benchmarked architectures, with the gap widening as RF increases, as expected. This effect is most pronounced for the CNNs, where the higher RF values in hls4ml, especially under the resource-focused strategy, cause substantial latency penalties due to multiplier reuse in convolution loops. The exception is the Jet architecture, where lower complexity and reduced per-layer parameter count potentially allow hls4ml to maintain a latency lower than SNL, with minimal variation across parameters.

Overall, the results, while constrained to our benchmark's limited set of models and synthesis configurations, demonstrate that SNL offers promising latency advantages for most tested architectures, often at the cost of higher BRAM and FF usage. For the FCNNs in particular, SNL also achieves notable DSP and LUT savings compared to hls4ml, especially at higher precisions. By contrast, hls4ml currently provides finer-grained control over resource–latency trade-offs through adjustable parameters beyond precision.

\section{Conclusion} 
\label{sec:conclusion}
Deploying neural networks in high-rate experimental environments demands low latency, efficient resource usage, and streamlined design flows. SNL is designed to meet these challenges by providing a specialized ML-to-FPGA framework with a workflow tailored for fast iteration. Through dynamic weight and bias reloading, SNL reduces the need for repeated synthesis when models change, enabling rapid adaptation to evolving experimental conditions.

Auto-SNL is a key addition we presented in this work that bridges Python-based model definition and hardware synthesis by automatically translating network architectures into HLS code and running the necessary implementation steps. This allows domain experts to deploy and refine models without delving into low-level C++ design, greatly shortening development cycles.

To evaluate the framework along with the full workflow, we performed a benchmark comparison against hls4ml across multiple neural network architectures, fixed-point precisions, and synthesis configurations targeting a Xilinx ZCU102 FPGA. The results showed that SNL achieves competitive or superior latency in most tested architectures, while in some cases also offering FPGA resource savings. At the same time, it is clear that SNL currently lacks an extensive parameter space outside of input and weight precision, limiting user control over the latency and resource utilization trade-offs. Additionally, the observed trends, while clear, are restricted to our non-exhaustive benchmark set and might not generalize to other models and deployment targets.

Looking forward, we plan to investigate a broader range of models and synthesis parameters, and target additional FPGA boards, expanding our current benchmark and further confirming the advantages SNL showcased in this work. Future work will also benefit from running the synthesized models on actual hardware, measuring on-chip inference latency and throughput to complement the results from post-synthesis reports. Finally, parallel to SNL's development, we aim to update Auto-SNL, exposing additional synthesis parameters and potentially offering users a graphical interface to streamline synthesis configuration. These developments will position SNL as a versatile yet accessible framework for real-time embedded neural network deployment, particularly in high-rate, resource-constrained environments.

\section*{Acknowledgment}
Abhilasha Dave, Larry Ruckman, James J. Russell, and Ryan Herbst's work was supported by the U.S. Department of Energy, under contract number DE-AC02-76SF0051. We would also like to extend our thanks to CMC Microsystems for access to Xilinx software licenses, as well as 3IT's GRAMS group for generously providing the hardware that helped make this work possible.

\bibliographystyle{IEEEtran}
\bibliography{main.bib}

\end{document}